\documentclass[journal]{IEEEtran}
\usepackage{amsmath,amsfonts,amssymb}
\usepackage{algorithmic}
\usepackage{algorithm}
\usepackage{array}
\usepackage[caption=false,font=normalsize,labelfont=sf,textfont=sf]{subfig}
\usepackage{textcomp}
\usepackage{stfloats}
\usepackage{url}
\usepackage{verbatim}
\usepackage{graphicx}
\usepackage{cite}

\usepackage{amssymb}
\usepackage{bbding}

\usepackage{hyperref}
\hypersetup{
colorlinks=true,
linkcolor=red, 
}
\usepackage{multirow}
\usepackage{booktabs}
\begin{document}

\title{Dynamic Patch-aware Enrichment Transformer for Occluded Person Re-Identification}

\author{Xin~Zhang,~~Keren~Fu,~~and~Qijun~Zhao

\IEEEcompsocitemizethanks{\IEEEcompsocthanksitem 
\IEEEcompsocthanksitem Xin Zhang is with the National Key Laboratory of Fundamental Science on Synthetic Vision, Sichuan University, Chengdu 610065, China. (E-mail: zhangxinchina1314@gmail.com)
\IEEEcompsocthanksitem Keren Fu and Qijun Zhao are
with the College of Computer Science, and
the National Key Laboratory of Fundamental Science on Synthetic Vision,
Sichuan University, Chengdu 610065, China. (E-mail: fkrsuper@scu.edu.cn; qjzhao@scu.edu.cn). \textit{(Corresponding author: Keren Fu).}\protect
}
}

\maketitle
\begin{abstract}
Person re-identification (re-ID) continues to pose a significant challenge, particularly in scenarios involving occlusions. Prior approaches aimed at tackling occlusions have predominantly focused on aligning physical body features through the utilization of external semantic cues. However, these methods tend to be intricate and susceptible to noise.
To address the aforementioned challenges, we present an innovative end-to-end solution known as the Dynamic Patch-aware Enrichment Transformer (DPEFormer). This model effectively distinguishes human body information from occlusions automatically and dynamically, eliminating the need for external detectors or precise image alignment.
Specifically, we introduce a dynamic patch token selection module (DPSM). DPSM utilizes a label-guided proxy token as an intermediary to identify informative occlusion-free tokens. These tokens are then selected for deriving subsequent local part features.
To facilitate the seamless integration of global classification features with the finely detailed local features selected by DPSM, we introduce a novel feature blending module (FBM). FBM enhances feature representation through the complementary nature of information and the exploitation of part diversity.
Furthermore, to ensure that DPSM and the entire DPEFormer can effectively learn with only identity labels, we also propose a Realistic Occlusion Augmentation (ROA) strategy. This strategy leverages the recent advances in the Segment Anything Model (SAM) \cite{kirillov2023segment}. As a result, it generates occlusion images that closely resemble real-world occlusions, greatly enhancing the subsequent contrastive learning process.
Experiments on occluded and holistic re-ID benchmarks signify a substantial advancement of DPEFormer over existing state-of-the-art approaches. The code will be made publicly available.
\end{abstract}


\section{Introduction}
\IEEEPARstart{P}{ERSON} 
re-identification (re-ID) is a challenging task involving the recognition and tracking of individuals across multiple non-overlapping cameras. Over recent years, it has attracted substantial research interest owing to its wide-ranging applications in various video surveillance scenarios. Thanks to the continuous advancements in deep learning techniques and the accessibility of extensive datasets, the field of re-ID has witnessed remarkable progress.
Numerous techniques~\cite{sun2018beyond,dai2019video,liu2022optimizing,jiang2022visible,liu2023deeptransformer,zhuang2023optimizing,dai2021generalizable,hu2021adversarial} have emerged to tackle complex issues like viewpoint variations and variable lighting conditions in the domain of person re-identification. However, most of these methods heavily rely on extensive datasets containing comprehensive and unobstructed pedestrian images for training deep neural networks. Consequently, their effectiveness may diminish when faced with occlusion scenarios, where individuals are occluded by objects like poles, vehicles, or walls. These occlusions introduce formidable challenges in precisely identifying individuals. As a result, occluded person re-identification emerges as a crucial area that deserves more in-depth exploration and investigation.
\begin{figure}
	\centering
	\includegraphics[width=9cm]{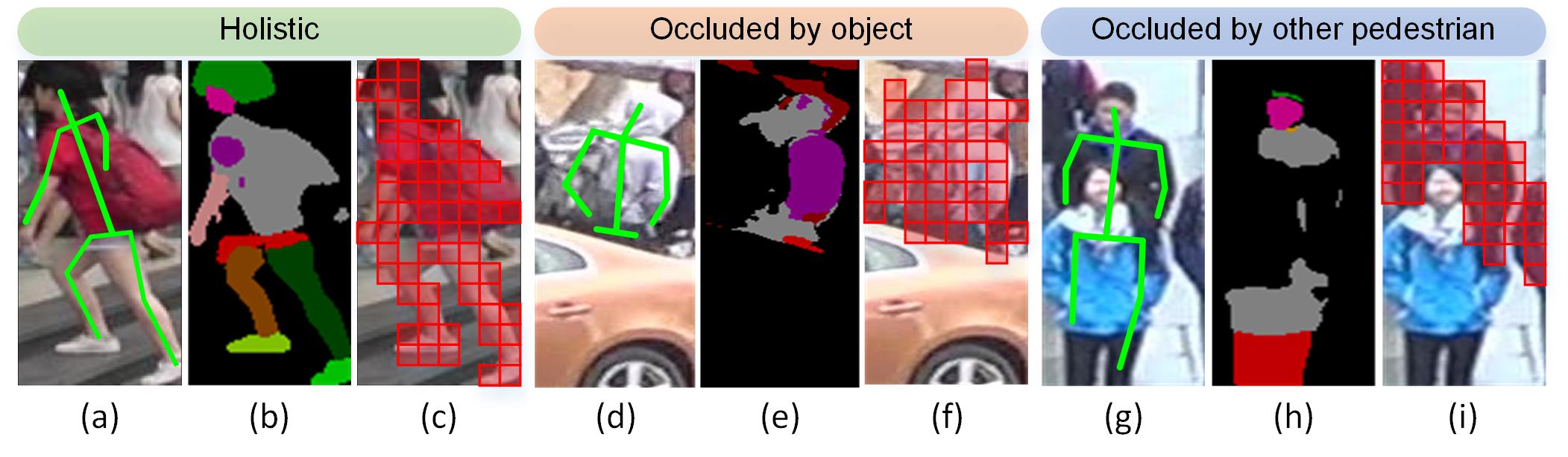}
        \vspace{-0.8cm}
        \caption{Applying pose estimation model \cite{sun2019deep} ((a), (d), (g)) and human parsing model \cite{li2020self} ((b), (e), (h)) to extract body information. Both models perform well when presented with holistic and object-occluded images but diminish when dealing with multi-pedestrian images. By contrast, our DPEFormer selects more accurate patches corresponding to body region ((c), (f), (i)).}
	\label{fig: illustration of intro}       
\end{figure}

In the context of occluded re-ID, a significant challenge arises from the presence of occluded regions, which tend to introduce noise and potentially lead to mismatches in the identification process. The crux of this challenge lies in the effective extraction of discriminative features from these non-occluded regions. Prevailing strategies to address this issue typically revolve around harnessing local features derived from distinct human body parts. Generally, these strategies rely on external cues provided by either semantic parsing~\cite{huang2020human,dou2023humanparsing1,somers2023humanparsing2} or pose estimation~\cite{miao2019pose,wang2022pose,YANG2023pose}.
In such approaches, a pre-trained pose or semantic detector is employed to identify landmarks or regions of interest within the images. These detected landmarks or regions serve as valuable cues to pinpoint the non-occluded areas and facilitate the alignment of local features during the learning process.
However, such solutions come with certain limitations. For instance, when dealing with considerable cross-domain disparities between training and testing data or in cluttered environments, the accuracy of off-the-shelf external detectors can be compromised. As depicted in Fig.~\ref{fig: illustration of intro}, in scenarios involving multi-pedestrian occlusions, pose estimation results might erroneously align with other pedestrians, resulting in inaccurate information extraction. Furthermore, human parsing models may not always recognize items carried by individuals, such as backpacks, hats, umbrellas, etc., potentially leading to the omission of crucial information for re-ID purposes. Additionally, the use of external detectors can introduce extra computational costs, which may be less advantageous in real-time surveillance applications.

Considering the preceding discussions, it is a common consensus that part-based representations hold promise as solutions to the challenges encountered in occluded person re-ID. Regrettably, the absence of part labels in such scenarios makes it challenging to train reliable intra-domain detectors. To address this predicament within the context of occluded person re-ID, this paper introduces a novel end-to-end learning model known as the Dynamic Patch-aware Enrichment Transformer (DPEFormer).
DPEFormer has the primary goal of localizing and selecting discriminative human body parts solely based on identity labels, without relying on external detectors. Leveraging the Transformer architecture~\cite{dosovitskiy2020image}, the pedestrian image is divided into smaller patches (\emph{e.g.}, Fig.~\ref{fig: illustration of intro} (c), (f), and (i)), each corresponding to a token. It is intuitive to expect that high-level tokens within the Transformer, associated with the same person, should exhibit more semantic similarity compared to tokens linked to occluded or background areas. Taking inspiration from this observation, we introduce the dynamic patch token selection module (DPSM).
DPSM operates by considering the information encapsulated within each patch token and assessing its significance through similarity calculations with a label-guided proxy token. These similarity values are then used to dynamically select the most crucial tokens via a first-order derivative process. It is worth noting that DPSM essentially functions as a hard attention mechanism, as it assigns binary weights (\{0,1\}) to tokens, in contrast to the soft attention mechanism that employs continuous weights ([0,1]). As depicted in Fig.~\ref{fig: illustration of intro}, DPEFormer, when trained with DPSM, demonstrates an improved ability to select more accurate patches for subsequent representation learning.

Furthermore, to enhance the aggregation of global features and part features selected by DPSM, we introduce a novel Feature Blending Module (FBM). FBM utilizes cross-attention mechanisms to comprehensively integrate both feature types, resulting in a more enriched final feature representation for re-ID tasks. Additionally, to facilitate improved learning for DPSM and the entire DPEFormer framework, we propose Realistic Occlusion Augmentation (ROA) at the image level. ROA takes advantage of recent advancements such as the Segment Anything Model (SAM)~\cite{kirillov2023segment}, allowing us to synthesize diverse and realistic occlusion data. This data includes scenarios involving multiple pedestrians and various objects, closely resembling real-world occlusion situations.
By training with ROA-augmented data, we can observe an improvement in the robustness of both DPSM and DPEFormer when it comes to handling occlusions. Note that ROA is introduced as an auxiliary training strategy and does not play a role in the inference phase. This design ensures that DPEFormer remains highly flexible during real-world applications.

In summary, to cope with occluded person re-ID, this paper presents three distinct contributing components, including Dynamic Patch Token Selection Module (DPSM), Feature Blending Module (FBM), and Realistic Occlusion Augmentation (ROA). The primary contributions can be succinctly summarized as follows:
\begin{itemize}
    \item We propose a patch-aware feature selection paradigm called DPSM for occluded person re-ID. The primary objective of DPSM is to pinpoint crucial human body patch tokens from the multitude of available tokens with occlusion/background. 
    \item Expanding upon our patch-aware feature selection paradigm, we introduce a Feature Blending Module (FBM) that enhances the synergy between global and carefully selected local features. This augmentation results in effective feature fusion.
    \item We propose Realistic Occlusion Augmentation (ROA) utilizing the Segment Anything Model (SAM)~\cite{kirillov2023segment}. ROA serves the dual purpose of reducing information redundancy during image augmentation while faithfully emulating realistic occlusion scenarios, including variations in contour details.    
\end{itemize}

The remainder of the paper is organized as follows. Section~\ref{sec: related work} discusses related work on holistic and occluded re-ID. Section~\ref{sec: method} provides detailed descriptions of the proposed DPEFormer framework, as well as key components: Dynamic Patch Selection Module (DPSM), Feature Blending Module (FBM), and Realistic Occlusion Augmentation (ROA). Experimental results, performance evaluations, and comparative analyses are included in Section~\ref{sec: Experiments}. Finally, conclusions are drawn in Section~\ref{sec: conclusion}.

\section{Related Work}
\label{sec: related work}
In this section, we give a brief review of existing methods for holistic person re-ID and occluded person re-ID.
\subsection{Holistic Person Re-Identification}
Person re-identification is a task that searches for or identifies a target person from multiple camera views. Existing methods can be roughly categorized into traditional~\cite{yang2014salient,liao2015person} and deep learning~\cite{chen2017beyond,zheng2019pyramidal} approaches. As a representative work, Yang \emph{et al.}~\cite{yang2014salient} propose a unique color descriptor and generate feature representation in the color space. With the emergence of large-scale datasets together with modern GPUs, deep learning techniques have been extensively adopted in the person re-ID field, among which part-based methods have shown competitive performance by leveraging fine-grained information of a human body. 
Sun \emph{et al.}~\cite{sun2018beyond} presents a simple and efficient part-based baseline convolutional network that employs a uniform partition strategy to learn part-level features. 
Wang \emph{et al.}~\cite{wang2018learning} design a multi-branch deep network, consisting of one global branch and two local branches with varying numbers of segmentation parts. Lin \emph{et al.}~\cite{lin2019improving} introduce additional attribute information, such as gender, hair length, age, backpacks, and so on, to enforce model learning of local details. 
Attention mechanisms have been adopted in \cite{tay2019aanet,chen2019mixed,zhang2022multi} to emphasize feature extraction on human body areas. 
\cite{zheng2021group,dai2021idm} fully exploit the knowledge from the source domain to address cross-domain re-ID.
Tan \emph{et al.}~\cite{He2023ISTA} incorporate spatial and temporal dual attention to refine pseudo-labels in unsupervised re-ID tasks.
Xu \emph{et al.}~\cite{Xu2023CASTOR} introduce a novel recycling strategy for pseudo-labels, addressing both pre-clustering and post-clustering stages.
Unfortunately, these above methods exhibit limited accuracy on retrieving individuals under occlusions, thereby hindering their applicability in crowded and complex scenarios.

\subsection{Occluded Person Re-Identification}
The current mainstream treatment for such occluded person re-ID is to adopt external information, including human parsing and pose estimation. 
Huang \emph{et al.}~\cite{huang2020human} are the pioneers in employing human parsing techniques for localizing body parts. 
He \emph{et al.}~\cite{miao2019pose} introduce a new network called Pose-Guided Feature Alignment (PGFA), which leverages pose landmarks to extract meaningful information while excluding occlusion noise. 
He \emph{et al.}~\cite{he2019foreground} introduce a new model called Foreground-aware Pyramid Reconstruction (FPR), which is dedicated to extracting features from foreground human body parts and computing matching scores between occluded pedestrians.
Gao \emph{et al.}~\cite{gao2020pose} propose an integrated framework called Pose-guided Visible Part Matching (PVPM) to learn discriminative features using a part visibility predictor and a pose-guided attention module. Wang \emph{et al.}~\cite{wang2020high} present an adaptive-direction graph convolutional (ADGC) model to learn semantic features, together with a cross-graph embedded alignment (CGEA) method for robust feature alignment. 
Yan \emph{et al.}~\cite{yan2023PRE-Net} propose a lightweight PRE to exploit local feature correlations and aggregate them without external detectors.

By leveraging transformer architecture, Jia \emph{et al.}~\cite{jia2022learning} propose DRL-Net  through disentangled representation learning.
Additionally, Wang \emph{et al.}~\cite{wang2022feature} propose a novel approach, called Feature Erasing and Diffusion Network (FED), which can effectively eliminate occlusion features in images.
To summarize, the above methods represent significant advances in the field of occluded person re-ID, and demonstrate encouraging performance for this challenging task.

Unlike the aforementioned previous methods that rely heavily on external human detectors or predefined occlusion priors from training sets, the proposed DPEFormer does not require any additional knowledge and also enables end-to-end training. 
Through careful consideration and design for occlusion elimination, DPEFormer demonstrates good generalizability in  handling different occlusion scenarios, and can also adapt to unseen occlusion cases effectively. 

\section{Methodology}
\label{sec: method}
\begin{figure*}[!t]
	\centering
        \hfil
	\includegraphics[width=17cm]{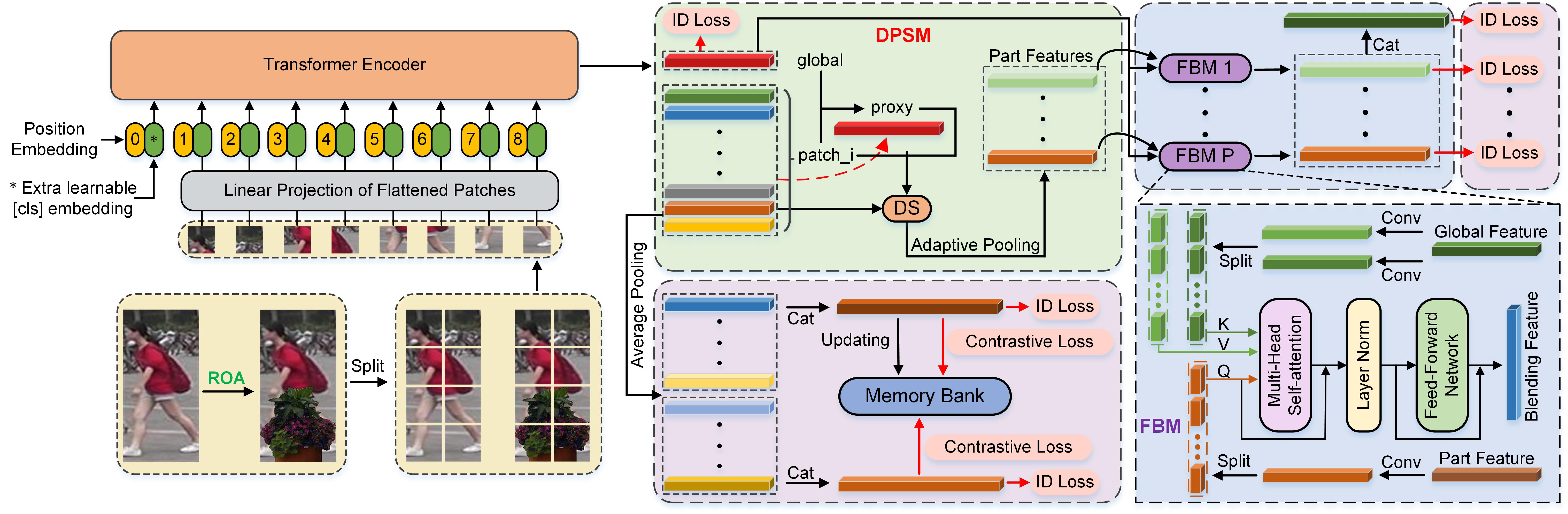}
	\caption{Framework of the proposed DPEFormer, which consists of four components: dynamic patch token selection module (DPSM), feature blending modules (FBMs), memory bank with contrastive loss, and realistic occlusion augmentation (ROA). Note that DPSM, FBM and ROA are three distinct contributions of this paper, which characterize the proposed DPEFormer.}
	\label{fig: network}       
\end{figure*}

The overall architecture of our model is illustrated in Fig.~\ref{fig: network}, which begins with pedestrian images as inputs. Note that these images will be augmented by ROA described later. 
Following previous works~\cite{li2021diverse,jia2022learning,wang2022feature}, the Vision Transformer (ViT)~\cite{dosovitskiy2020image} is adopted as our feature extractor. A learnable $[cls]$ classification token and position embeddings are prepended to the input image. After the transformer encoder, the output feature can be expressed as $f\in \mathbb{R}^{(N+1)\times c}$, where $N+1$ including one $[cls]$ token and $N$ image patch tokens, and $c$ indicates the dimension. In practice, $N$ and $c$ are 128 and 768 (the same as ViT-Base~\cite{dosovitskiy2020image}), respectively. Then, we feed the output feature $f$ to DPSM to perform token selection, which discards a certain amount of tokens while retaining the rest for subsequent feature representation. After DPSM, several local part features from adaptive pooling of the selected tokens, as well as the global classification token, are further fed to FBMs to embed global classification features into part features to obtain enriched feature representation. After that, the outputs of all FBMs (corresponding to individual parts) are concatenated to form the final pedestrian descriptor, which is supervised by identity loss. Meanwhile, we employ the memory bank and contrastive loss proposed in \cite{ge2020self} for enhanced supervision, as shown by the pink region in Fig.~\ref{fig: network}. This part works concurrently with the DPSM and FBMs during training, but takes tokens prior to selection as inputs.  Details of each component will be introduced in the following sections.

\subsection{Dynamic Patch Token Selection Module (DPSM)}\label{sec: DPSM}
Recognizing that not all patch tokens from ViT contribute equally to the representation of a pedestrian, especially an occluded one, we propose a dynamic patch token selection module (DPSM) to exclude a certain amount of ineffective patches. In turn, DPSM aims to emphasize contributions of those advantageous ones based on specific rules to enhance model performance. 
Firstly, we introduce a proxy token $f_{proxy}$ that is carefully selected from all patch tokens using a similarity measure that compares with the global features, namely the class token. This proxy serves as a key link between the global and patch features and will be used for subsequent patch token selection to alleviate the impact of occlusion. 
The features obtained after ViT backbone can be denoted as $f = \{f_g,f_1,f_2,...,f_N\}$, where $f_g$ is the class token and can be deemed as the global features for the re-ID task.
Owing to that all the token features are normalized after ViT, we directly measure similarities by computing the inner product between $f_g$ and each patch token $f_i$, where $i=1,2,...,N$. 
Next, we identify the patch token having the greatest similarity with the class token $f_g$ as the proxy token $f_{proxy}$. Mathematically, the lower-index $proxy$ of $f_{proxy}$ can be formulated as:
\begin{equation}
    proxy=\underset{i}{\operatorname{arg\,max}}\, (f_g f_1^{\intercal},f_g f_2^{\intercal},\dots, f_g f_N^{\intercal}).
\end{equation}
\par
Since $f_{proxy}$ is obtained from the strongest patch token representation and it is reasonable to assume it corresponds to the pedestrian body area, $f_{proxy}$ can be considered as a reliable feature representation. To further evaluate similarities between $f_{proxy}$ and the other tokens, we perform dot product again between $f_{proxy}$ and each patch embedding, obtaining a set of similarity scores as:
\begin{equation}
    \mathtt{S}^p=\{f_{proxy} f_1^{\intercal},f_{proxy} f_2^{\intercal},\dots, f_{proxy} f_N^{\intercal}\},
\end{equation}
where we note that we do not yet exclude $f_{proxy}$ itself from the set, yielding $N$ scores in total, among which the maximum score equals 1.

\begin{figure}
	\centering
	\includegraphics[width=8.5cm]{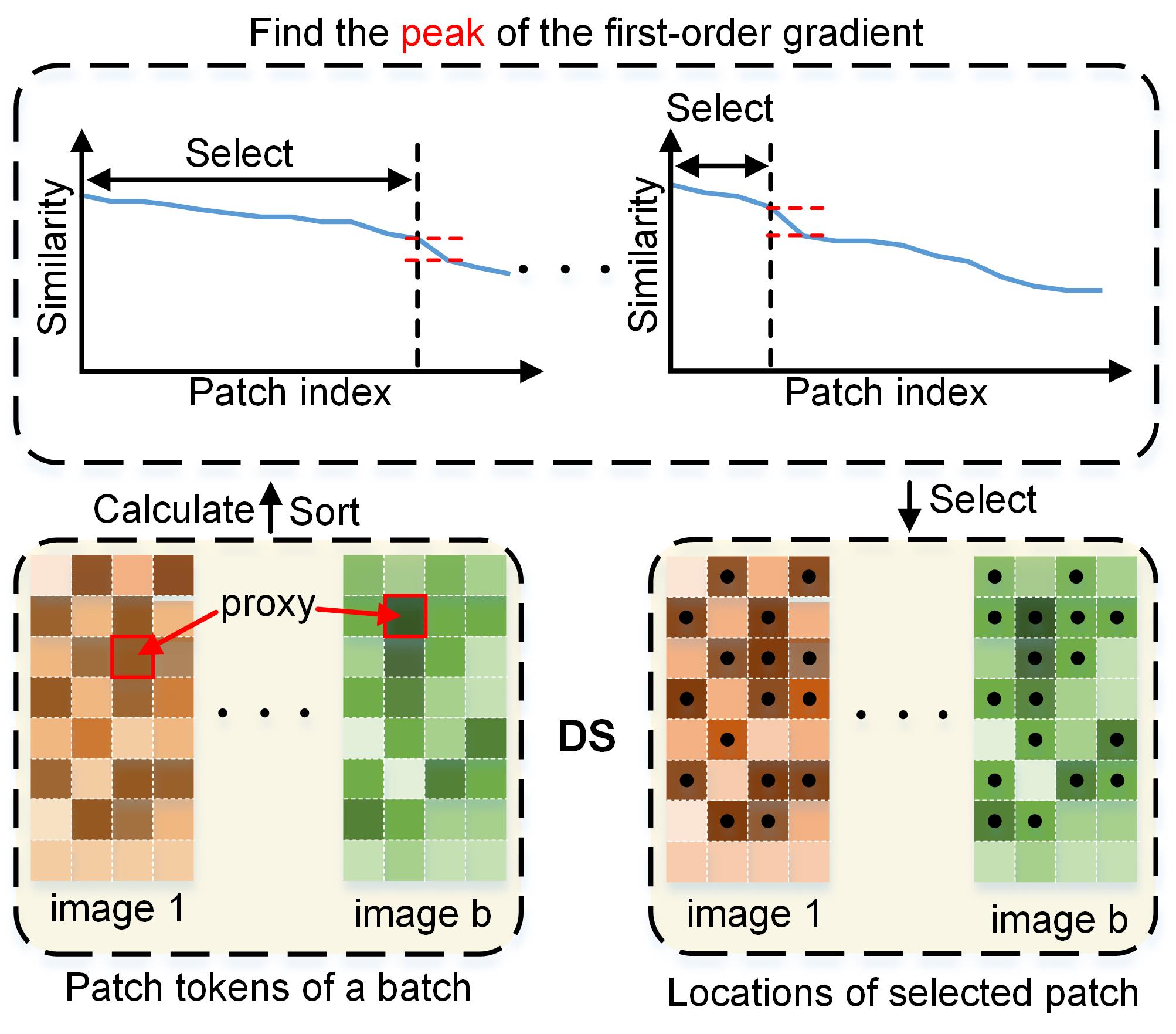}
	\caption{Illustration of the dynamic selection (DS) process for patch tokens, where notation ``b'' refers to the batch size. To perform DS, we calculate the similarity scores between the proxy token with all patch tokens, and such scores are sorted in descending order and serve as support for selecting essential tokens.}
	\label{fig: Dynamic-PSM}       
\end{figure}

As shown in Fig.~\ref{fig: Dynamic-PSM}, we then sort all similarity values in descending order via $\operatorname{Sort}(\cdot)$ operation as:
\begin{equation}
    \mathtt{S}^o=\operatorname{Sort}(\mathtt{S}^p),
\end{equation}
and calculate the first-order gradient of $\mathtt{S}^o$:
\begin{equation}
    \mathtt{D}[i]=\mathtt{S}^o[i]-\mathtt{S}^o[i+1],\\
\end{equation}
and let $\mathtt{Diff}$ be denoted as: 
\begin{equation}
    \mathtt{Diff}=\{\mathtt{D}[1],\mathtt{D}[2],\dots,\mathtt{D}[N-1]\},
\end{equation}
where $N$ means the total number of patch tokens.
We assume that the pedestrian body and other information, such as background and occlusion, belong to distinct categories in the feature space. Our goal is to divide the patch tokens into two clusters, among which one corresponds to body information and the other is associated with background/occlusion information. 
To address this, we refer to $\mathtt{Diff}$ and consider the position with the maximal gradient magnitude, namely the maximal first-order difference, as the splitting point. The underlying assumption is that there should be a distinct feature transition when body features change to those tokens that are contaminated by occlusions or other interfering information. Therefore, this splitting point is chosen as: 
\begin{equation}
    k =\underset{i}{\operatorname{arg\,max}} (\mathtt{Diff}).
\end{equation}
This means those tokens corresponding to the first $k$ scores in $\mathtt{S}^o$ will be selected as body feature tokens to represent a pedestrian, and the rest tokens are deemed less effective and will be discarded subsequently. 

Despite the above maximal gradient magnitude demonstrating certain efficacy (will be discussed in Section~\ref{sec: Experiments}), we also observe some failure cases in practice due to the strong assumption of a clear boundary between body tokens and background/occlusion tokens. If a pedestrian presents a similar appearance to the occlusion object or background, this strategy may lead to the misclassification of similar parts as disturbance, resulting in the selection of too few patch tokens.
Consequently, this reduction in the number of selected patch tokens can compromise the model's ability to accurately capture comprehensive pedestrian information, ultimately affecting its overall performance.

To address this issue, we set up an initial minimum for $k$, denoted as $k_{min}$, to define the least number for patch selection. The value of $k_{min}$ is empirically determined (see Section~\ref{sec: ablation studies}) and acts as a relaxation condition.
Finally, the dynamic number of the selected candidate tokens can be defined as $k \leftarrow max(k,k_{min})$. Lastly, we perform adaptive average pooling on the selected $k$ patch tokens, yielding $P$ part features denoted as $f_{{part}_i}$, where $i= 1,\dots, P$, and together with $f_g$ as the outputs of DPSM.

\textbf{Discussion.} 
Actually, the proposed hard attention mechanism DPSM can be deemed as performing a two-class clustering process to divide the tokens into two classes, and the proposed algorithm is a simple tailored one to achieve such a goal. Given the clustering guidance by $f_{proxy}$, DPSM treats $f_{proxy}$ as a seed and compares all the samples with this seed. Tokens similar to $f_{proxy}$ are clustered into one class, while those far away are put into the other class. It is also worth noting that, in the view of clustering, one can also apply any clustering algorithm, like naive K-means, to bi-partition tokens $\{f_1,f_2,...,f_N\}$. However, such a way is much less efficient as we need to handle enormous training images during learning. The clustering results from K-means may also be less stable and reasonable. Therefore, we reject those clustering approaches and employ a proxy token-based sharpest gradient scheme to establish the splitting point.        

\begin{figure}
	\centering
	\includegraphics[width=8.5cm]{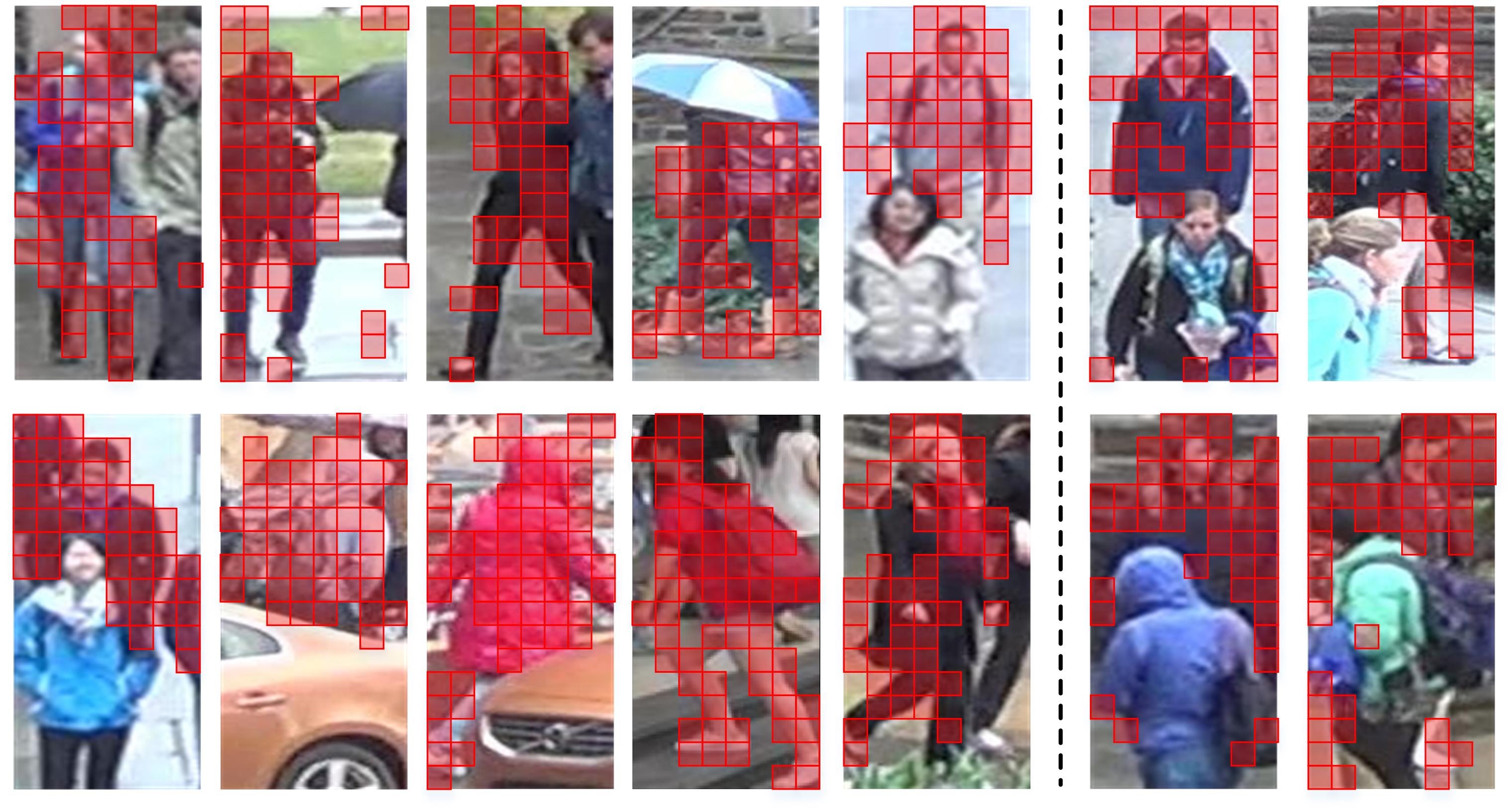}
        \caption{Visualization of selected patches by DPSM. Examples on the left of the dash line show that the majority of selected patches exhibit alignment with pedestrian bodies, whereas examples on the right of the line show some visually failed cases where the selected patches do not align well with the desired regions.}
	\label{fig: select patch}       
\end{figure}

To visualize the selected patch tokens, we map the selected ones back to image space, where raw image patches corresponding to the selected patch tokens are highlighted. The visualization results are shown in Fig.~\ref{fig: illustration of intro} and Fig.~\ref{fig: select patch}. One can observe that the majority of selected patches align with pedestrian bodies, effectively avoiding occlusions caused by other pedestrians or objects. However, DPSM sometimes fails to ``perfectly'' select patches. Instead, the selected patches could cover the background or some occlusion areas. 
One possible reason is that the global self-attention mechanism of Transformer has exchanged information across all patch tokens, making their feature representation complicated and not well aligned with the original image space, \emph{e.g.}, a patch token corresponds to the background of the original image may also convey pedestrian identity representation. Despite the above explanation, we observe notable performance gains after using DPSM.

\subsection{Feature Blending Module (FBM)}
\label{sec: FBM}
To accurately identify pedestrians, it is imperative to consider both contextual information from the global features $f_g\in \mathbb{R}^{1\times 768}$ and detailed information from the part features $f_{{part}_i}\in \mathbb{R}^{1\times 768}, i= 1,\dots, P$.
In recent work~\cite{wang2022interact}, a module was developed specifically for communicating information between global and part features in the multi-modal re-ID task. Inspired by \cite{wang2022interact}, we propose a feature enrichment strategy for occluded re-ID.
To this end, given these two types of features, we embed global context information into each local part feature to enrich representation power via FBM. Fig.~\ref{fig: network} bottom right illustrates the structure of FBM, which we describe below using the $i$-th part features $f_{{part}_i}$ and the global features $f_g$ for simplicity.

The underlying idea of FBM is to leverage the well-known multi-head self-attention (MHSA) mechanism as an enrichment mean (Fig.~\ref{fig: network}), rather than simple point-to-point addition (i.e., $f_{{part}_i}+f_g$) or concatenation. We initially implement three $1\times 1$ convolution layers on global and part features to obtain $f_g^{'}$, $f_g^{''}$ and $f_{{part}_i}^{'}$.
Then we split these feature vectors into 48 non-overlapping sub-parts, with each sub-part having a fixed size of 16. Such 48 sub-parts are then formulated as regular inputs to the attention mechanism.  
Learnable positional embeddings are added to these feature sub-parts. In contrast to the standard MHSA, we modify MHSA to derive query set $Q_{p}$ from the feature sub-parts of $f_{{part}_i}^{'}$,
and to derive key set $K_{g}$ and value set $V_{g}$ from the feature sub-parts of $f_g^{'}$ and $f_g^{''}$. The graphic illustration is shown in Fig.~\ref{fig: network} bottom right, and our modified MHSA enables comprehensive mutual interactions between the part and global features, instead of naive point-to-point integration performed by simple addition. 

Given $Q_{p},K_{g},V_{g}$, the modified MHSA is formulated as:
\begin{equation}
    \operatorname{MHSA}_{g \rightarrow p}(Q_p,K_g,V_g)=\operatorname{Concat}(head_0,\dots,head_h)W^O,
\end{equation}
where $W^O$ is the output transformation matrix for integrating multi-head outputs, and $head_i$ is computed as:
\begin{equation}
    head_i=\operatorname{Attention}\left(Q_p^i, K_g^i, V_g^i\right),
\end{equation}
where $Q_p^i, K_g^i, V_g^i$ are the query, key, and value matrices of the $i$-th head, respectively.
The general computation of Attention($\cdot$) is defined as:
\begin{equation}
    \operatorname{Attention}(Q, K, V)=\operatorname{Softmax}\left(\frac{Q K^T}{\sqrt{d_k}}\right) V,
\end{equation}
where $\sqrt{d_k}$ denotes a scaling factor, and is equal to $4(\sqrt{16})$ in our case.
So generally, the feature blending process in FBM is represented as:
\begin{equation}
f_{{mhsa}_i}\!=\!\operatorname{RE}\left(\operatorname{MHSA}_{g \rightarrow p}\left(\operatorname{LN}\left(\operatorname{FPR}\left(f_g^{'},f_g^{''}, f_{{part}_i}^{'}\right)\!+\!\mathrm{PE}\right)\right)\right),
\end{equation}
where $\operatorname{RE}$ represents the reshape operation, which aims to maintain the same dimension as $f_{{part}_i}$. $\operatorname{PE}$ and $\operatorname{LN}(\cdot)$ denote position embedding and layer normalization, respectively. $\operatorname{FPR}(\cdot)$ represents feature partition and sub-part matrix formulation, namely transforming $\mathbb{R}^{1\times 768}$ into $\mathbb{R}^{48\times 16}$.
Finally, the entire forward blending process of FBM is computed as:
\begin{equation}        
f_{{fbm}_i}=\operatorname{LN}\left(f_{{part}_i}+f_{{mhsa}_i}\right)+\operatorname{FFN}\left(\operatorname{LN}\left(f_{{part}_i}+f_{{mhsa}_i}\right)\right),
\end{equation}
where $\operatorname{FFN}(\cdot)$ denotes the feed-forward layer, and $f_{{fbm}_i}$ denotes the obtained blending features by embedding $f_g$ into $f_{{part}_i}$.
\par
To ensure semantic diversity, we employ FBM for each part, namely, there will be $P$ FBM in total. Finally, all enriched part features are concatenated to form the final pedestrian descriptor:
\begin{equation}
    f_{final}=\operatorname{Concat}(f_{{fbm}_i},\dots,f_{{fbm}_P}),
\end{equation}
where $P$ denotes the total number of parts. During the training process, $f_g$ is employed solely for classification and discarded in the final feature representation. The final representation $f_{final}\in \mathbb{R}^{1\times 3072}$ encompasses diverse detail-specific information from different local parts, and is also enriched by global knowledge through the proposed feature embedding via FBM.

\subsection{Realistic Occlusion Augmentation (ROA)}
\label{sec: ROA} 
A significant challenge faced by existing approaches in occluded person re-ID lies in the limited availability of occlusion data. In response to this challenge, researchers have explored various strategies. Zhong \emph{et al.}~\cite{zhong2020random} introduced a method known as Random Erasing (RE), which involves the random occlusion of a rectangular region with random pixel values.
Chen \emph{et al.}~\cite{chen2021occlude} adopted a random cut-and-paste approach, where a patch is cropped from a training image and pasted onto the input image. Following the idea of \cite{chen2021occlude}, Wang \emph{et al.}~\cite{wang2022feature} further cropped occlusion objects like bags, manually from the training set and pasted them. 
Such an approach can be deemed as incorporating additional prior information specifically related to pedestrian occlusion.
The aforementioned approaches have made significant progress in addressing data scarcity. However, a common limitation shared by these methods is that the synthesized images appear to be coarse and not realistic. To overcome this limitation, we present Realistic Occlusion Augmentation (ROA) that significantly diminishes data redundancy and greatly enhances the realism of generated images, better simulating real-world occlusions. 

\begin{figure}[t]
\centering
\includegraphics[width=8.5cm]{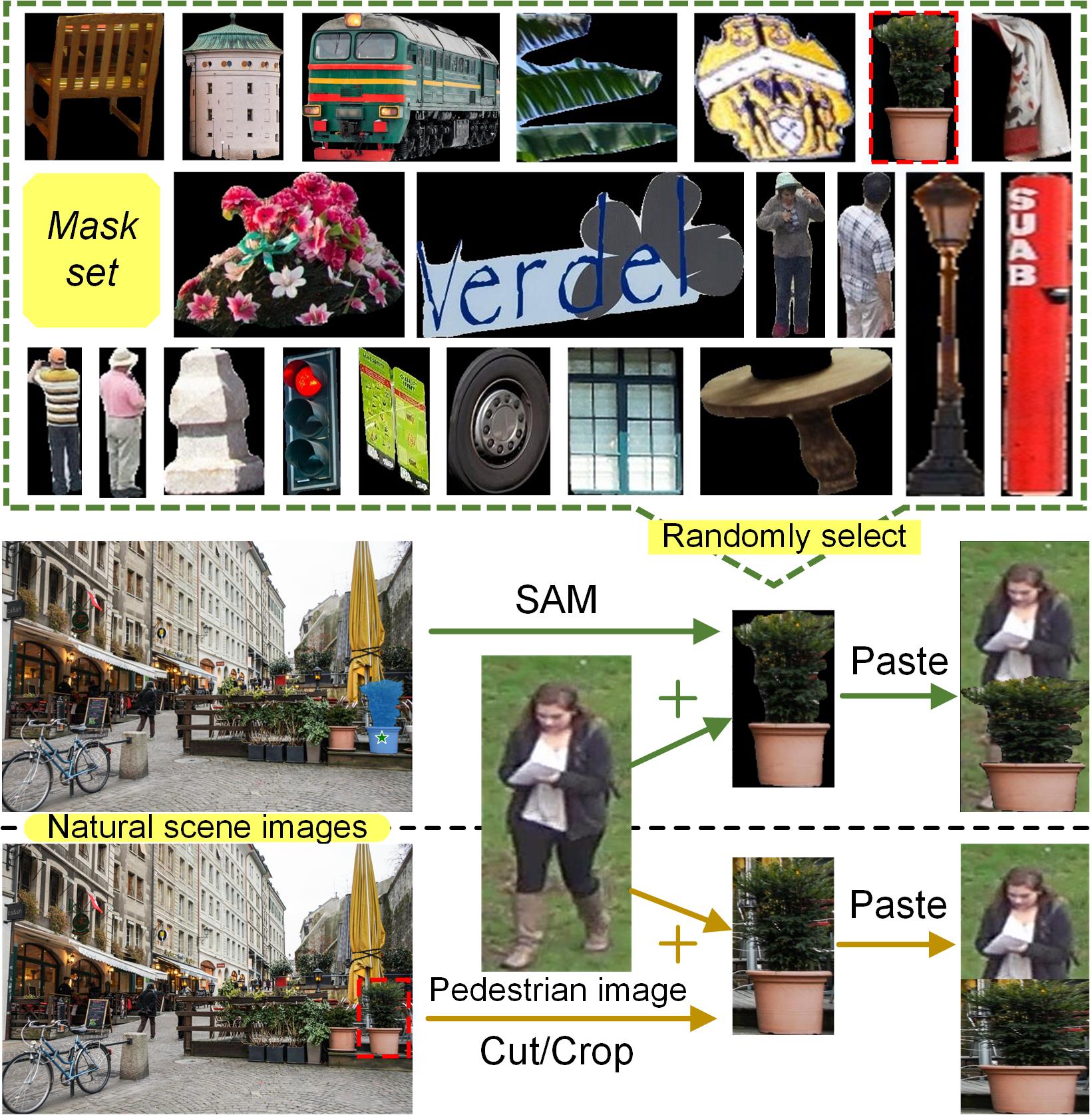} 
\caption{Comparison between the proposed ROA (upper) and existing Cut\&Paste (bottom). For ROA, we pre-generate a mask set using SAM~\cite{kirillov2023segment}, from which we randomly select masks for occlusion synthesis during training. Detailed algorithm can be found in Algorithm~\ref{alg:algorithm}.}
\label{fig: data augmentation}
\end{figure}

The procedure of the proposed ROA is outlined in Algorithm \ref{alg:algorithm}, whose mask set is generated by the Segment Anything Model (SAM)~\cite{kirillov2023segment} applied on natural images (Fig.~\ref{fig: data augmentation}).
In Algorithm~\ref{alg:algorithm}, variable $random\_area$ means the area of the bounding box of an occlusion object, and therefore, the actual occlusion area is usually smaller than this value.
Following the steps in Algorithm~\ref{alg:algorithm}, ROA is able to generate occluded images that closely resemble real-world scenarios, with occlusions manifesting in different directions and covering various regions of a pedestrian body. 
As shown in Fig.~\ref{fig: data augmentation}, the generated occlusion image by ROA is more natural and exhibits diverse contour details when compared to cut\&paste~\cite{chen2021occlude}.

\begin{algorithm}[tb]
\caption{Realistic Occlusion Augmentation (ROA)}
\label{alg:algorithm}
\renewcommand{\algorithmicrequire}{\textbf{Input:}}
\renewcommand{\algorithmicensure}{\textbf{Output:}}
\begin{algorithmic}[1] 
\REQUIRE Training set: $\mathcal{X}_{train}$; Mask set: $\mathcal{X}_{mask}$ (generated by SAM~\cite{kirillov2023segment}, detailed in Sec. \ref{sec:experiment setting}).
\ENSURE Realistic Occlusion Augmentation set $\mathcal{X}_{roa}$.
\STATE \textit{$\%$For simplicity, we use the width and height of the bounding box to represent the size of an object mask.}
\FOR{each mini-batch $\mathcal{B} \subset \mathcal{X}_{train}$}
\FOR{each $\textbf{x}_i \in \mathcal{B}$}
\STATE Randomly select a mask $\textbf{x}_{mask}$ from $\mathcal{X}_{mask}$.
\STATE Obtain the size $mask\_h\times mask\_w$ of $\textbf{x}_{mask}$ and $image\_h\times image\_w$ of $\textbf{x}_{i}$, and the area $area$ of $\textbf{x}_{i}$.
\STATE Select a random value $random\_area$ between $1/2*area$ and $3/4*area$.
\IF{$mask\_h/mask\_w > 2$}
\STATE $resize\_w=random\_area/image\_h$
\STATE $resize\_h=image\_h$
\ELSE
\STATE $resize\_w=image\_w$
\STATE $resize\_h=random\_area/image\_w$
\ENDIF
\STATE Scale $\textbf{x}_{mask}$ to size $resize\_w\times resize\_h$
\STATE Perform random horizontal flip for $\textbf{x}_{mask}$.

\STATE Randomly place $\textbf{x}_{mask}$ at one of the three positions $\{(0,0),(0,image\_h-resize\_h),(image\_w-resize\_w,0)\}$
\STATE Generate  $\textbf{x}_{roa}$ by occluding $\textbf{x}_{i}$ with $ \textbf{x}_{mask}$.
\STATE Add $\textbf{x}_{roa}$ to $\mathcal{X}_{roa}$.
\ENDFOR
\ENDFOR
\STATE \textbf{return} $\mathcal{X}_{roa}$
\end{algorithmic}
\end{algorithm}

\subsection{Loss Function}\label{sec: Loss Function}
For better training our DPEFormer model, we adopt the same memory bank strategy as in \cite{ge2020self} to regularize feature learning. Since the learning process of DPSM may not be stable at the beginning (patch tokens reside in learning and have not yet converged), the patch selection process yields less reliable results. So we apply the memory bank to the patch tokens prior to DPSM and construct the associated contrastive loss and identity loss.  

\subsubsection*{Memory Bank}

Following \cite{ge2020self}, the memory initialization takes place only once at the beginning of the training. Moreover, to obtain identity centers, all extracted features from the training set are averaged and stored in a memory-based feature dictionary. Memory updating is performed at every forward inference stage. As shown by Fig.~\ref{fig: network}, the feature vector of each training instance is utilized to update the corresponding dictionary vector during forward computation. For an instance with identity $j$, after a series of operations (the ViT encoder, average pooling, and concatenation operation, yielding features $f_{q_j}$), the updating process is formulated as:
\begin{equation}
d_{j}\leftarrow \mu \cdot d_{j} + (1 - \mu) \cdot f_{q_j},
\label{eq:update}
\end{equation}
where $\mu$ is a momentum factor, and $d_j$ is the dictionary vector corresponding to identity $j$. Note that to the maintain stability of dictionary updating, we only use non-ROA samples to update the dictionary. Meanwhile, the memory bank is adopted for calculating \emph{contrastive loss}.

\subsubsection*{Contrastive Loss}
During training, we compare the training instance's features $f_{q_j}$ to all the dictionary vectors $\{d_1,d_1,$
$\dots,d_{M}\}$ ($M$ means the number of identity classes) using the InfoNCE loss \cite{oord2018representation}:
\begin{equation}\label{eq:loss}
\mathcal{L}_{con} = -\log\frac{exp(f_{q_j} \cdot d_j^{\intercal}/\tau)}{\sum_{l=1}^{M}{exp(f_{q_j} \cdot d_l^{\intercal}/\tau)}},
\end{equation}
where $\tau$ is a temperature factor, $d_j$ is the positive centroid feature vector for instance $f_{q_j}$.
The loss enforces an instance close to its positive centroid $d_j$ and deviates away from the other negative ones. Note that in practice, we render ROA images $\mathcal{X}_{roa}$ with a smaller weight (about 0.3) on this contrastive loss term while the original images $\mathcal{X}_{train}$ are rendered weight 1.

\subsubsection*{Total Loss}
We choose the cross-entropy loss as the identity loss $\mathcal{L}_{id}$ to train the model in our experiments, and all the global and part features are under the constraint of $\mathcal{L}_{id}$. The overall classification loss $\mathcal{L}_{cls}$ then can be formulated as: 

\begin{equation}\label{eq: cls loss}
\begin{split}
\mathcal{L}_{cls}&=\sum^P_{i=1}(\mathcal{L}_{id}(\Theta(f_{{fbm}_i}),y))+\mathcal{L}_{id}(\Theta(f_{q_j}),y)\\
       &+\mathcal{L}_{id}(\Theta(f_g),y)+\mathcal{L}_{id}(\Theta(f_{final}),y),
\end{split}
\end{equation}
where $\Theta$ represents the probability prediction function, which comprises a bottleneck layer followed by a fully connected layer. $y$ denotes the ground truth identity label. $P$ is the total number of parts. Finally, the total loss $L_{total}$ is defined as:
\begin{equation}\label{eq: total loss}
\mathcal{L}_{total} = \mathcal{L}_{con}+\mathcal{L}_{cls}.
\end{equation}

\section{Experiments and Results}
\label{sec: Experiments}
\subsection{Datasets and Evaluation Settings}
\textbf{Occluded-DukeMTMC}~\cite{miao2019pose} represents one of the most challenging occluded person ReID datasets due to the diversity of scenes and distractions encountered. It comprises 15,618 training images of 702 persons, 2,210 query images of 519 persons, and 17,661 gallery images of 1,110 persons. 

\textbf{Occluded-REID}~\cite{zhuo2018occluded} focuses on occluded individuals captured through mobile cameras, comprising 2,000 images spanning 200 unique identities. Within each identity, the dataset includes five unobstructed full-body images and five images featuring severe occlusions, showcasing diverse viewpoints and occlusion types.

\textbf{Market-1501}~\cite{zheng2015scalable} is one of the most well-known holistic person ReID datasets, with 12,936 training images of 751 persons, 19,732 gallery images, and 3,368 query images of 750 persons captured from six cameras. Few of images in this dataset are occluded.

\textbf{DukeMTMC}~\cite{zheng2017unlabeled} consists of 16,522 training images of 702 persons, 2,228 queries of 702 persons, and 17,661 gallery images of 702 persons. The images are captured by eight different cameras, making it more challenging. As it contains more holistic images than occluded ones, this dataset can be treated as a holistic ReID dataset.

\subsection{Experimental Settings}\label{sec:experiment setting}
\paragraph{Experimental details}
During training, each pedestrian image is resized to a resolution of $256\times 128$. 
For each iteration, we select a batch of 64 images, consisting of 16 identities, each with 4 distinct images. The Adam optimizer is employed for model optimization, with the weight decay factor set to $1\times 10^{-4}$. 
We used the initial 100 images from the SA-1B dataset~\cite{kirillov2023segment} as the original images to extract the masks. By utilizing SAM, we generated a collection $\mathcal{X}_{mask}$ of 9,913 occlusion masks.
The minimum threshold of selected patch tokens is set to 48. The momentum factor $\mu$ is set to 0.2 in Eq.~\eqref{eq:update}. 
Following \cite{wang2022feature}, the part number $P$ is set to 4.
To achieve optimal performance, we train the model for 120 epochs, initializing the learning rate of the parameters to $1\times 10^{-4}$, adopting cosine learning rate decay. 
The platform for the implementation of our experiment is PyTorch~\cite{paszke2019pytorch}. All global and local features are subject to identity loss. Lastly, only four enriched part features are concatenated to produce a final 3072-dimensional person descriptor for testing.

For a fair comparison, our backbone network just adopts ViT~\cite{dosovitskiy2020image} without employing a specific sliding window or other specialized configurations. 

\paragraph{Protocols} The model is evaluated with two standard metrics: Rank-1 accuracy and mean average precision (mAP). All experiments are carried out in a single query mode.

\subsection{Comparison with State-of-the-Art Methods}
We compare our method with state-of-the-art methods on both occluded and holistic re-ID datasets in Table~\ref{tab:Occluded} and Table~\ref{tab: comparison on holistic dataset}, respectively. The backbones of compared methods are ResNet-50~\cite{he2016deep} and Vision Transformer~\cite{dosovitskiy2020image}.

\hphantom\noindent\textbf{Comparisons on Occluded Re-ID Datasets.}
DPEFormer is evaluated with recent state-of-the-art occluded re-ID methods. The comparison results are presented in Table~\ref{tab:Occluded}.
We can see that PAT~\cite{li2021diverse} significantly improves accuracy by adopting a transformer encoder-decoder structure capable of leveraging diverse part-aware attention maps. These attention maps serve as specific part detectors and boost network performance.
Our DPEFormer achieves 69.9\% Rank-1 accuracy, 82.8\% Rank-5 accuracy, 86.4\% Rank-10 accuracy, and 58.9\% mAP, which surpasses all other types of occluded methods by a substantial margin. Notably, our method does not rely on external cues. 
It is designed to mitigate interference of occlusion solely based on the available training data and the designed architecture. 

\hphantom\noindent\textbf{Comparisons on Holistic Re-ID Datasets.}
We also evaluate our model on two holistic person Re-ID datasets and compare it with other state-of-the-art methods in Table~\ref{tab: comparison on holistic dataset}.
While our proposed strategies and modules aim to address occlusion problems, they may not fully function well for holistic scenarios. Nonetheless, we still achieve an impressive Rank-1=95.4\%/mAP=88.1\% and Rank-1=90.0\%/mAP=80.3\% accuracy on Market-1501 and DukeMTMC, respectively, surpassing all other types of holistic re-ID methods.
This comprehensive performance indicates that our proposed methods effectively exploit a robust feature representation not only for occlusion challenges but also for holistic ones.

\begin{table}[t]
    \centering
    \caption{Performance comparison with state-of-the-art methods on occluded datasets Occlude-DukeMTMC and Occluded-REID (\%). ``*'' denotes that external information is utilized. The best results are shown in \textbf{bold}.}
    \label{tab:Occluded}
    \scalebox{0.725}{
    \begin{tabular}{ l|c|cc|cc}
        \toprule[1pt]
        \multirow{2}{*}{Backbone}&\multirow{2}{*}{Methods} & \multicolumn{2}{c|}{Occlude-DukeMTMC} & \multicolumn{2}{c}{Occluded-REID} \\
        \cline{3-6}
        & & Rank-1 & mAP & Rank-1 & mAP \\
        \hline
        \multirow{13}{*}{CNN} 
        &Part Aligned \cite{zhao2017deeply} (ICCV 17) & 28.8 & 20.2 & - & - \\
        &HACNN \cite{li2018harmonious} (CVPR 18) & 34.4 & 26.0 & - & - \\
        &Part Bilinear* \cite{suh2018part} (ECCV 18) & 36.9 & - & - & - \\
        &FD-GAN* \cite{ge2018fd} (NIPS 18) & 40.8 & - & - & - \\
        &DSR \cite{he2018deep} (CVPR 18) & 40.8 & 30.4 & 72.8 & 62.8 \\
        &SFR \cite{he2018recognizing} (ArXiv 18) & 42.3 & 32.0 & - & - \\
        &PCB \cite{sun2018beyond} (ECCV 18) & 42.6 & 33.7 & 41.3 & 38.9 \\
        &Adver Occluded \cite{huang2018adversarially} (CVPR 18) & 44.5 & 32.2 & - & - \\
        &PVPM* \cite{gao2020pose} (CVPR 20) & 47.0 & 37.7 & 70.4 & 61.2 \\
        &PGFA* \cite{miao2019pose} (ICCV 19) & 51.4 & 37.3 & - & - \\ 
        &HOReID* \cite{wang2020high} (CVPR 20) & 55.1 & 43.8 & 80.3 & 70.2 \\
        &MoS \cite{jia2021matching} (AAAI 21) & 61.0 & 49.2 & - & - \\
        &OAMN \cite{chen2021occlude} (ICCV 21) & 62.6 & 46.1 & - & - \\
        &MMNet \cite{Tu2022MMNet} (TITS 22) & 56.1 & 50.1 & - & - \\
        &PRE-Net \cite{yan2023PRE-Net} (TCSVT 23) & 67.1 & 54.3 & - & - \\
        \hline        
        \multirow{4}{*}{Transformer}
        &PAT \cite{li2021diverse} (CVPR 21) & 64.5 & 53.6 & 81.6 & 72.1 \\
        &DRL-Net \cite{jia2022learning} (TMM 23) & 65.0 & 50.8 & - & - \\ 
        &FED \cite{wang2022feature} (CVPR 22) & 68.1 & 56.4 & 86.3 & 79.3 \\
        &\textbf{DPEFormer} \emph{(Ours)} & \textbf{69.9} & \textbf{58.9} & \textbf{87.0} & \textbf{79.5} \\ 
        \bottomrule[1pt]
    \end{tabular}}
\end{table}

\begin{table}[t]
    \centering
    \caption{Performance comparison with state-of-the-art methods on holistic datasets Market-1501 and DukeMTMC (\%).}
    \label{tab: comparison on holistic dataset}
    \scalebox{0.87}{
    \begin{tabular}{l| c |c c|c c}
        \toprule[1pt]
        \multirow{2}{*}{Backbone}&\multirow{2}{*}{Methods} & \multicolumn{2}{c|}{Market-1501} & \multicolumn{2}{c}{DukeMTMC} \\
        \cline{3-6}
         & & Rank-1 & mAP & Rank-1 & mAP \\
        \hline
        \multirow{13}{*}{CNN}
        &DSR \cite{he2018deep} (CVPR 18) & 83.6 & 64.3 & - & - \\
        &PSE* \cite{Sarfraz2018APE} (CVPR 18) & 87.7 & 69.0 & 27.3 & 30.2 \\
        &VCFL \cite{Liu2019ViewCF} (ICCV 19) & 89.3 & 74.5 & - & - \\
        &PGFA* \cite{miao2019pose} (ICCV 19) & 91.2 & 76.8 & 82.6 & 65.5 \\
        &MVPM \cite{SunCYX19} (ICCV 19) & 91.4 & 80.5 & 83.4 & 70.0 \\
        &PCB \cite{sun2018beyond} (ECCV 18) & 92.3 & 77.4 & 81.8 & 66.1 \\
		&OAMN* \cite{chen2021occlude} (ICCV 21) & 92.3 & 79.8 & 86.3 & 72.6\\
        &VPM \cite{sun2019deep} (CVPR 19) & 93.0 & 80.8 & 83.6 & 72.6 \\
        &SFT \cite{Luo2019SpectralFT} (ICCV 19) & 93.4 & 82.7 & 86.9 & 73.2 \\
        &AANet* \cite{tay2019aanet} (CVPR 19) & 93.9 & 82.5 & 86.4 & 72.6 \\
        &Circle \cite{sun2020circle} (CVPR 20) & 94.2 & 84.9 & - & - \\
        &HOReID* \cite{wang2020high} (CVPR 20) & 94.2 & 84.9 & 86.9 & 75.6 \\
        &PRE-Net \cite{yan2023PRE-Net} (TCSVT 23) & 94.5 & 86.0 & 88.9 & 76.5 \\
        \hline
        \multirow{4}{*}{Transformer}
        &DRL-Net \cite{jia2022learning} (TMM 23) & 94.7 & 86.9 & 88.1 & 76.6 \\
        &FED \cite{wang2022feature} (CVPR 22) & 95.0 & 86.3 & 89.4 & 78.0\\
        &PAT \cite{li2021diverse} (CVPR 21) & 95.4 & 88.0 & 88.8 & 78.2 \\
        &\textbf{DPEFormer} \emph{(Ours)} & \textbf{95.4} & \textbf{88.1} & \textbf{90.0} & \textbf{80.3} \\
        \bottomrule[1pt]
    \end{tabular}}
\end{table}

\begin{table}[t]
	\centering
	\caption{Performance analysis of each component in DPEFormer on Occluded-DukeMTMC.}
 	\label{tab: Ablation of each component}
	\begin{tabular}{c|cccc|cc}
	    \toprule
		Index & RE & ROA & DPSM & FBM &  Rank-1 & mAP \\
		\hline
		1 & - & - & - & -  & 60.4 & 50.4 \\
		2 & \Checkmark & - & - & - & 60.5 & 53.0  \\
		3 & - & \Checkmark & - & - & 67.1 & 58.1\\
		4 & - & \Checkmark & \Checkmark & - & 68.7 & 58.7  \\
		5 & - & \Checkmark & - & \Checkmark & 68.1 & 58.6 \\
		6 & - & \Checkmark & \Checkmark & \Checkmark & \textbf{69.9} & \textbf{58.9} \\
		\bottomrule[1pt]
	\end{tabular}
\end{table}

\subsection{Ablation Study}
\label{sec: ablation studies}
In this section, we present a comprehensive series of ablation studies to thoroughly analyze our proposed framework.

\hphantom\noindent\textbf{Effectiveness of each component.}
In Table~\ref{tab: Ablation of each component}, we present a series of ablation studies that investigate the individual contributions of various components: random erasing (RE)~\cite{zhong2020random}, realistic occlusion augmentation (ROA), the dynamic patch token selection module (DPSM), and the feature blending module (FBM). In detail,
``index 1" corresponds to the baseline model, where different images of the same pedestrian are used as input for contrastive learning. Subsequently, we incrementally introduce augmentations and modules, denoted as indexes from 1 to 5: baseline + RE, baseline + ROA, baseline + ROA + DPSM, baseline + ROA + FBM, and DPEFormer, respectively.
Comparing ``index 0" (the baseline) to ``index 1/2" (baseline + RE and baseline + ROA), we observe that the inclusion of data augmentation strategies, specifically ROA, significantly enhances performance. This improvement can be attributed to the generation of a more diverse and realistic set of augmented images, which in turn aids the model during training.

Comparing ``index 3'' to ``index 4'', we observe that the DPSM leads to a further enhancement in representations, resulting in 1.6\% improvement in Rank-1 accuracy. In Fig.~\ref{fig: illustration of intro} and \ref{fig: select patch}, we provide a visual representation of the patches selected by DPSM, demonstrating its effectiveness in directing the model's focus towards extracting human body information while filtering out tokens associated with occlusions or background elements.
Furthermore, when comparing ``index 3'' to ``index 5'', the FBM contributes to performance gains, showing improvements of 1.0\% in Rank-1 accuracy and 0.5\% in mAP. This underscores FBM's ability to diversify local part features and enhance local representations by effectively fusing complementary global information.
Ultimately, ``index 6'' (DPEFormer) achieves the highest accuracy, highlighting the efficacy of each component, both individually and in synergy.

\hphantom\noindent\textbf{Parameter analysis of DPSM.}
We investigate the impact of the patch token selection module on model performance. To this end, we introduce a fixed threshold selection strategy referred to as Fixed Threshold PSM (FPSM). In FPSM, we modify the token selection process in the Dynamic Patch Selection Module (DPSM) to a fixed number while keeping all other training strategies unchanged.
Table~\ref{tab: fixed number} presents our results. We observe an obvious trend as the number of selections increases: both mAP and Rank-1 accuracy improve. However, this trend changes after reaching a threshold of 48, where we begin to observe a gradual decrease in performance with further increases in the selection number.
This observation suggests that an abundance of patch tokens, which incorporate occlusion or background information, can degrade the feature representation. Conversely, an insufficient number of patch tokens may fail to capture the necessary information required for accurate person characterization.
Balancing the selection of patch tokens is a critical factor in achieving an optimal feature representation and robustness in person recognition. Therefore, it becomes essential to establish a minimum threshold for the number of patch token selections to ensure a more effective implementation of the dynamic selection strategy. Setting this threshold ensures that an adequate number of candidate patch tokens are retained, striking the right balance in the process.

\begin{table}[t]
    \centering
    \caption{Performance analysis of different fixed selection numbers in FPSM on Occluded-DukeMTMC. ``Number=128'' denotes that all the patch tokens are selected.}
    \label{tab: fixed number}
    \begin{tabular}{c|cccc}
        \toprule[1pt]
        Number & Rank-1 & Rank-5 & Rank-10 & mAP \\
        \hline
            12 & 68.1 & 82.5 & 86.2 & 58.3 \\
            24 & 68.2 & 81.7 & 86.4 & 58.4 \\
            48 & \textbf{69.0} & \textbf{82.8} & 86.5 & \textbf{58.7} \\
            96 & 68.4 & 82.2 & \textbf{86.6} & 58.7 \\  
            128 & 68.1 & 82.1 & 86.4 & 58.3 \\          
        \bottomrule[1pt]
    \end{tabular}
\end{table}

\begin{table}[t]
    \centering
    \caption{Parameter analysis for the minimum selection number $k_{min}$ on Occluded-DukeMTMC. ``$k_{min}=0$'' denotes that no minimum number of patch token selections is imposed. }
    \label{tab: min_number}
    \begin{tabular}{l|cccc}
    \toprule[1pt]
        $k_{min}$ & Rank-1 & Rank-5 & Rank-10 & mAP \\
        \hline
            0 & 69.3 & 82.5 & 86.4 & 58.8 \\    
            12 & 68.2 & 82.0 & 86.0 & 58.5\\        
            24 & 68.3& 82.3 & 86.2 & 58.5 \\
            36 & 68.5 & 82.4 & 86.4 & 58.6 \\
            48 \emph{(Ours)} & \textbf{69.9} & \textbf{82.8} & \textbf{86.6} & \textbf{58.9} \\
            60& 68.6 & 82.3 & 86.4 & 58.5 \\ 
            72& 68.4 & 82.2 & 86.2 & 58.4 \\ 
    \bottomrule[1pt]
    \end{tabular}
\end{table}

We conducted experiments to determine the optimal hyperparameter, denoted as $k_{min}$, by varying its values and assessing the impact on the model performance, as presented in Table~\ref{tab: min_number}. Notably, when there were no constraints on the minimum selection number ($k_{min}=0$), meaning tokens were selected solely based on their first gradient value, the model achieved competitive performance. As we increased $k_{min}$ from 12, we observed an improvement in Rank-1/mAP performance by 1.6\%/1.3\% (reaching $k_{min}=48$), indicating that this parameter became beneficial for learning discriminative features. However, further increases in $k_{min}$ led to performance degradation, as more noisy tokens, including occlusions and background information, were included.
Additionally, the dynamic selection strategy outperformed the fixed number strategy presented in Table~\ref{tab: fixed number}. This improvement can be attributed to the enhanced flexibility of the dynamic selection module, which adapts better to the varying number of characteristics in each image.
From another perspective, it is worth noting that the model achieved its best performance when the hyperparameter was set to 48, as evidenced in both Table~\ref{tab: fixed number} and Table~\ref{tab: min_number}.

Furthermore, Table~\ref{tab: comparison before or after DPSM} presents a comparison of recognition accuracy achieved by using the feature representations before and after the DPSM for inference, without the need for retraining. ``Before'' signifies the utilization of a pooling operation on all patch tokens to form the feature representation, while ``After'' denotes the use of the patch tokens selected by DPSM for pooling.
It is evident that utilizing the feature representation after the DPSM results in higher accuracy, underscoring the module's effectiveness.

\begin{table}[t]
    \centering
    \caption{Comparison between using features before and after the application of DPSM for pedestrian representation on Occluded-DukeMTMC.}
    \label{tab: comparison before or after DPSM}
    \begin{tabular}{ l|cccc}
        \toprule[1pt]
        Feature & Rank-1 & Rank-5 & Rank-10 & mAP \\
        \hline
        Before & 68.1 & 82.2 & 86.4 & 58.3 \\
        After & \textbf{68.7} & \textbf{82.4} & \textbf{86.4} & \textbf{58.7}\\
        \bottomrule[1pt]
    \end{tabular}
\end{table}

\hphantom\noindent\textbf{Necessity for the proxy token in DPSM.}
We refrain from using global features as proxies due to their inclusion of information from all body parts. The results, as shown in Table~\ref{tab: proxy}, clearly demonstrate that employing a selected patch token as the proxy yields superior performance compared to using global features (cls token). This highlights that the selected patch token provides more representative information for token selection. The effectiveness of using patch tokens as proxies can be attributed to the elimination of discrepancies caused by redundant information present in global features.

\begin{table}[t]
    \centering
    \caption{Comparison of using the global token or patch token as the proxy token on Occluded-DukeMTMC.}
    \label{tab: proxy}
    \begin{tabular}{ l|cccc}
        \toprule[1pt]
        Proxy token & Rank-1 & Rank-5 & Rank-10 & mAP \\
        \hline
        Global token & 68.4 & 81.9 & 86.4 & 58.6 \\
        Closest patch token \emph{(Ours)} & \textbf{69.9} & \textbf{82.8} & \textbf{86.6} & \textbf{58.9}\\
        \bottomrule[1pt]
    \end{tabular}
\end{table}

\hphantom\noindent\textbf{FBM \emph{vs.} REM.}
The efficacy of our FBM is confirmed through a comparative analysis with an alternative fusion strategy known as REM, introduced by Wang \emph{et al.}~\cite{wang2022interact}. All other experimental settings remain constant. The results, as presented in Table~\ref{tab: comparison of FBM and REM}, reveal that REM exhibits a decrease in Rank-1 accuracy by 1.2\% and a decline in mAP by 0.5\% when compared to our proposed FBM. This outcome underscores the superior capability of our FBM in effectively integrating and fusing global and local features within the context of DPEFormer.

\begin{table}[t]
    \centering
    \caption{Comparison of the proposed FBM with REM on Occluded-DukeMTMC.}
    \label{tab: comparison of FBM and REM}
    \begin{tabular}{ l|cccc}
        \toprule[1pt]
        Method & Rank-1 & Rank-5 & Rank-10 & mAP \\
        \hline
        REM \cite{wang2022interact} & 68.7 & 81.7 & 85.6 & 58.3 \\
        FBM \emph{(Ours)} & \textbf{69.9} & \textbf{82.8} & \textbf{86.6} & \textbf{58.9} \\ 
        \bottomrule[1pt]
    \end{tabular}
\end{table}

\hphantom\noindent\textbf{On Realistic Occlusion Augmentation.}
We assess the efficacy of our Realistic Occlusion Augmentation (ROA) approach through a comparative analysis with two established methods: Random Erasing (RE)~\cite{zhong2020random} and the Cut\&Paste method~\cite{chen2021occlude}.
Specifically, in the Cut\&Paste configuration, we employ the same object instances as ROA to ensure a fair comparison. In this scenario, we directly extract rectangular regions from scene images instead of employing our occlusion generation method.
The results of this evaluation are presented in Table~\ref{tab: comparison of ROA}.
We observed that the Cut\&Paste method outperforms Random Erasing, possibly due to its ability to provide rich occlusion information. However, ROA significantly outperforms both of these augmentation methods. This substantial improvement validates the effectiveness of ROA, which leverages the Segment Anything Model (SAM)~\cite{kirillov2023segment} to generate realistic occluded images. ROA proves to be a superior approach for augmenting occluded person images.

To ensure a fairer comparison, we replaced the occlusion augmentation in FED and the ROA in DPEFormer with the widely recognized occlusion augmentation RE. The results are also presented in Table~\ref{tab: comparison of ROA}. We can see that our approach consistently outperforms FED, underscoring the effectiveness of our meticulously designed modules (DPSM and FBM). 
It is worth mentioning that FED achieves competitive performance, especially when leveraging its NPO augmentation strategy, which incorporates prior manual occlusion information from the training set. 

Furthermore, to evaluate the influence of the generated occlusion mask set size on model performance, we employed varying quantities of original images choosen from SA-1B datasets. The findings, as illustrated in Table~\ref{tab: number of X_mask}, reveal a performance uptick with a candidate image count ranging from 50 to 100. Optimal performance is attained at approximately 100 candidates. Beyond this threshold, augmenting the number of images yields minimal impact. At this juncture, the model is exposed to a sufficiently diverse array of occlusion shapes, and texture saturation has reached an optimal value. Hence, our investigation establishes a candidate set size of 100 as the optimal enhancement size.

\begin{table}[t]
    \centering
    \caption{The influence of the quantity of selected images from the SA-1B dataset, utilized as source images for $\mathcal{X}_{mask}$ formation.}
    \label{tab: number of X_mask}
    \begin{tabular}{ l|cccc}
        \toprule[1pt]
        Number & Rank-1 & Rank-5 & Rank-10 & mAP \\
        \hline
        50 & 69.0 & 82.3 & 86.4 & 58.6 \\
        100 \emph{(Ours)} & \textbf{69.9} & \textbf{82.8} & \textbf{86.6} & \textbf{58.9} \\ 
        200 & 69.7 & 82.6 & 86.3 & 58.8 \\ 
        \bottomrule[1pt]
    \end{tabular}
\end{table}

\begin{table}[t]
    \centering
    \caption{Comparison of occlusion augmentation method on Occluded-DukeMTMC. }
    \label{tab: comparison of ROA}
    \begin{tabular}{l|l|cccc}
        \toprule[1pt]
        Augmentation & Model & Rank-1 & Rank-5 & Rank-10 & mAP \\
        \hline 
        ROA \emph{(Ours)} & \multirow{3}{*}{DPEFormer} & \textbf{69.9} & \textbf{82.8} & \textbf{86.6} & \textbf{58.9} \\
        Cut\&Paste \cite{chen2021occlude} &  & 66.9 & 79.5 & 84.1 & 56.0 \\
        RE \cite{zhong2020random}&  & 63.9 & 78.9 & 84.0 & 55.5 \\
        \hline
        RE \cite{zhong2020random} & \multirow{2}{*}{FED \cite{wang2022feature}} & 62.6 & 77.7 & 83.0 & 55.2 \\
        NPO \cite{wang2022feature} &  & 68.1 & - & - & 56.4 \\
        \bottomrule[1pt]
    \end{tabular}
\end{table}

\begin{figure}[ht]
	\centering
	\includegraphics[width=8.5cm]{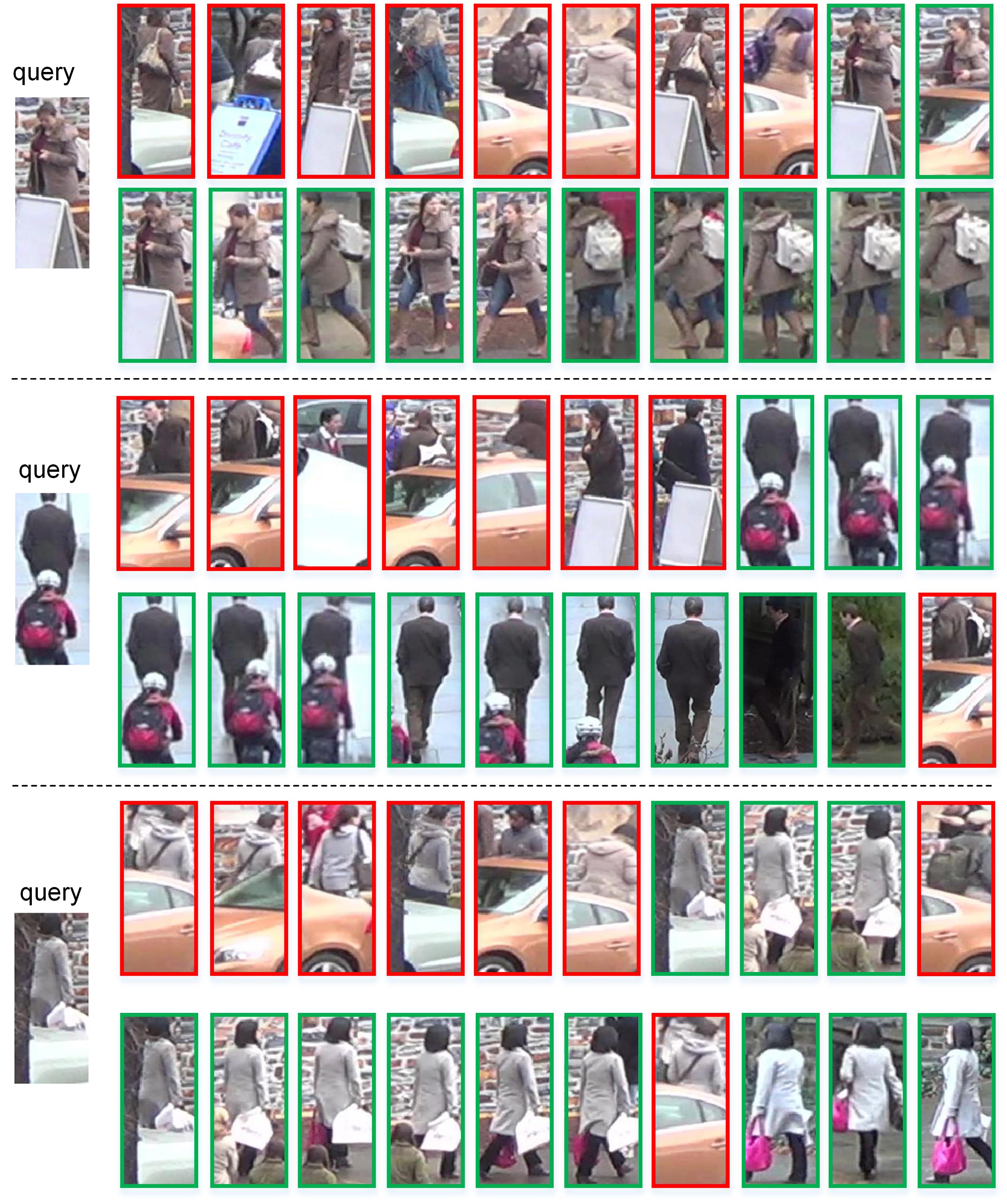}
        \caption{Retrieval results on Occluded-DukeMTMC. In each query group, the first and second rows display the top-10 matched images retrieved by the \emph{baseline} model and our proposed DPEFormer, respectively. Images enclosed within green rectangles represent correct identity matches to the queries, while those within red rectangles signify false matching results. It is important to note that, from left to right, the matching scores progressively decrease.}
	\label{fig: retrieval images}       
\end{figure}

\hphantom\noindent\textbf{About the contrastive loss.}
We introduced an additional contrastive loss after the Feature Blending Module (FBM). As outlined in Table~\ref{tab: contrast loss}, the results demonstrate that the inclusion of extra contrastive loss led to a decrease of 1.5\% in mAP and a 3.1\% reduction in Rank-1 accuracy.
Our analysis aligns with the observation that body region information differs between occluded and non-occluded images. Furthermore, this discrepancy is accentuated by the DPSM. Introducing contrastive loss as a constraint in this context may hinder the learning of crucial discriminative features, potentially impacting DPSM's performance. However, the contrastive loss applied to the memory bank serves as guidance for the Vision Transformer (ViT) backbone to extract features devoid of occlusions. This preparation enhances the subsequent processing by DPSM.

\begin{table}[t]
    \centering
    \caption{Analysis of the utilization of contrastive loss on Occluded-DukeMTMC. ``One" denotes the execution of contrastive loss solely after the backbone, whereas ``two" indicates additional implementation of contrastive loss post the FBM.}
    \label{tab: contrast loss}
    \begin{tabular}{ l|cccc}
        \toprule[1pt]
        Contrastive loss & Rank-1 & Rank-5 & Rank-10 & mAP \\
        \hline
        One \emph{(Ours)} & \textbf{69.9} & \textbf{82.8} & \textbf{86.6} & \textbf{58.9} \\ 
        Two & 66.8 & 81.3 & 85.9 & 57.4 \\
        \bottomrule[1pt]
    \end{tabular}
\end{table}

\subsection{Qualitative Analyses}
We present qualitative experimental results to demonstrate the efficacy of DPEFormer. In Fig.~\ref{fig: select patch}, we provide visualizations of the patch tokens selected by DPSM, mapping them back to their corresponding locations in the image space. We have highlighted the raw image patches that correspond to these selected patch tokens. These images depict instances where pedestrians and objects occlude the scene.
Notably, a significant proportion of the selected patches align precisely with the pedestrians' body regions, effectively mitigating occlusions caused by other pedestrians or objects. This visualization underscores the robustness and effectiveness of DPEFormer in handling occlusions within complex real-world scenarios.

Fig.~\ref{fig: retrieval images} illustrates how DPEFormer overcomes occlusions by presenting some retrieval results.
Each set comprises an occluded query person image displayed on the left, along with two rows of images on the right. These right-hand images represent the top-10 matches generated by the baseline model and our proposed DPEFormer.
Upon observation, it is evident that DPEFormer excels at overcoming occlusions, accurately identifying images of the same pedestrian. In contrast, the baseline network exhibits heightened sensitivity to occlusions, resulting in a substantial number of false-positive matches.

\section{Conclusion}
\label{sec: conclusion}
This paper introduces DPEFormer, a novel end-to-end architecture specifically tailored for occlusion re-identification tasks. DPEFormer operates at the patch token level, allowing for the automatic and insightful selection of human body part features free from occlusions, all without the need for additional supervision. Moreover, we present a novel Feature Blending Module meticulously crafted to enhance feature representation. It achieves this by harnessing the complementary nature of information and capitalizing on the diverse aspects of body parts. In addition, we introduce a Realistic Occlusion Augmentation (ROA) strategy, grounded in SAM, which enables the generation of more authentic occluded data. This augmentation significantly enriches DPEFormer's overall learning capability, making it highly adaptable to real-world scenarios.
Our extensive experiments and comparisons showcase substantial improvements achieved by DPEFormer over state-of-the-art models in the domain of occlusion handling. We believe that DPEFormer offers fresh insights into addressing occlusion challenges within the occluded re-ID problem, and we hope it could inspire further research.

\bibliographystyle{IEEEtran}
\bibliography{egbib}

\end{document}